\documentclass{article} 
\usepackage{iclr2026_conference,times}

\usepackage{amsmath,amsfonts,bm}

\def\eqref#1{equation~\ref{#1}}

\def\1{\bm{1}}

\DeclareMathAlphabet{\mathsfit}{\encodingdefault}{\sfdefault}{m}{sl}
\SetMathAlphabet{\mathsfit}{bold}{\encodingdefault}{\sfdefault}{bx}{n}

\usepackage{hyperref}
\usepackage{url}
\usepackage{booktabs}       
\usepackage{amsfonts}       
\usepackage{nicefrac}       
\usepackage{microtype}      
\usepackage{xcolor}         

\usepackage{multirow}
\usepackage{array}
\usepackage{colortbl}
\usepackage{graphicx}
\usepackage{tabularx}
\usepackage{booktabs}
\usepackage{subcaption}
\usepackage{floatrow}
\usepackage{natbib}
\usepackage{wrapfig}

\usepackage{amsmath}
\usepackage{algorithm}
\usepackage{algpseudocode}

\newcommand{\vcenteredhbox}[1]{\begingroup
  \setbox0=\hbox{#1}\parbox{\wd0}{\box0}\endgroup}
\newfloatcommand{capbtabbox}{table}[][\FBwidth]
\definecolor{lightpurple}{RGB}{230,230,250}
\definecolor{lightpink}{RGB}{255,228,225}
\definecolor{lineblue}{RGB}{93,144,191}
\definecolor{brickred}{rgb}{0.8, 0.25, 0.33}
\definecolor{brickblue}{rgb}{0.302, 0.475, 0.788}
\usepackage[dvipsnames]{xcolor}

\hypersetup{
  colorlinks=true,
  linkcolor=brickred,
  citecolor=brickred,
  urlcolor=brickred
}

\title{D-TPT: Dimensional Entropy Maximization for Calibrating Test-Time Prompt Tuning in Vision-Language Models}
\author{Jisu Han , Wonjun Hwang\thanks{Corresponding author.}\\School of Electrical Engineering\\ Korea University \\ \texttt{\{jisuhan,wjhwang\}@korea.ac.kr} \\ }
\iclrfinalcopy 
\begin{document}

\maketitle

\begin{abstract}
Test-time adaptation paradigm provides flexibility towards domain shifts by performing immediate adaptation on unlabeled target data from the source model. Vision-Language Models (VLMs) leverage their generalization capabilities for diverse downstream tasks, and test-time prompt tuning has emerged as a prominent solution for adapting VLMs. In this work, we explore contrastive VLMs and identify the modality gap caused by a single dominant feature dimension across modalities. We observe that the dominant dimensions in both text and image modalities exhibit high predictive sensitivity, and that constraining their influence can improve calibration error. Building on this insight, we propose dimensional entropy maximization that regularizes the distribution of textual features toward uniformity to mitigate the dependency of dominant dimensions. Our method alleviates the degradation of calibration performance in test-time prompt tuning, offering a simple yet effective solution to enhance the reliability of VLMs in real-world deployment scenarios.
\end{abstract}

\section{Introduction}
Foundation models provide generalized performance through massive data~\citep{rombach2022high,oquab2023dinov2,kirillov2023segment}; among them, Vision Language Models (VLMs) such as CLIP~\citep{radford2021clip} are applied to diverse downstream tasks~\citep{zhang2021tip,goyal2023finetune}. Based on the observation that the zero-shot performance of VLMs variates considerably depending on prompt configuration, prompt tuning methods that optimize the prompt to determine more appropriate ones for the target task through training~\citep{zhou2022coop,zhou2022cocoop}. Prompt tuning-based approaches are applied across various areas, including continual learning~\citep{l2p,wang2022dualprompt}, test-time adaptation~\citep{niu2024test,zhang2024dpcore}, and visual recognition tasks~\citep{zhou2023zegclip,he2023unsupervised}, which utilize pretrained models.

\begin{wrapfigure}{r}{0.3\textwidth}
    \centering
    \vspace{-7mm}
    \includegraphics[width=0.95\columnwidth]{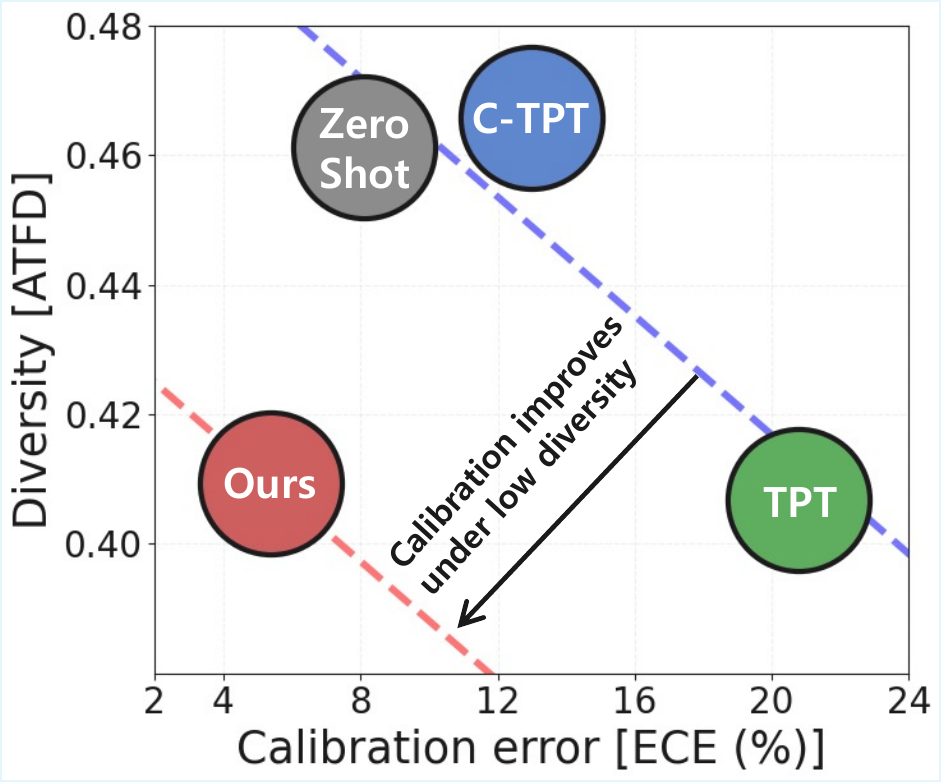}
    \caption{Correlation between text diversity and calibration.}
    \label{fig:ece_atfd}
    \vspace{-3.4mm}
\end{wrapfigure}

Furthermore, Test-time Prompt Tuning (TPT) extends its applicability to unlabeled settings by adapting prompts online~\citep{shu2022tpt}. Although TPT effectively improves predictive accuracy, it simultaneously induces overconfidence due to its reliance on prediction entropy minimization as the objective function. Recent studies~\citep{zhang2024come,han2025ranked} attempt to mitigate the degradation in model calibration performance caused by entropy minimization and improve practical applicability. Notably, calibration methods in test-time adaptation for VLMs are based on empirical observations. While increasing text feature diversity effectively improves calibration~\citep{yoon2024ctpt,sharifdeen2025otpt}; however diversity-based approaches suffer from a lack of understanding regarding its underlying mechanism.

\begin{figure}[t]
    \centering
    \includegraphics[width=1\linewidth]{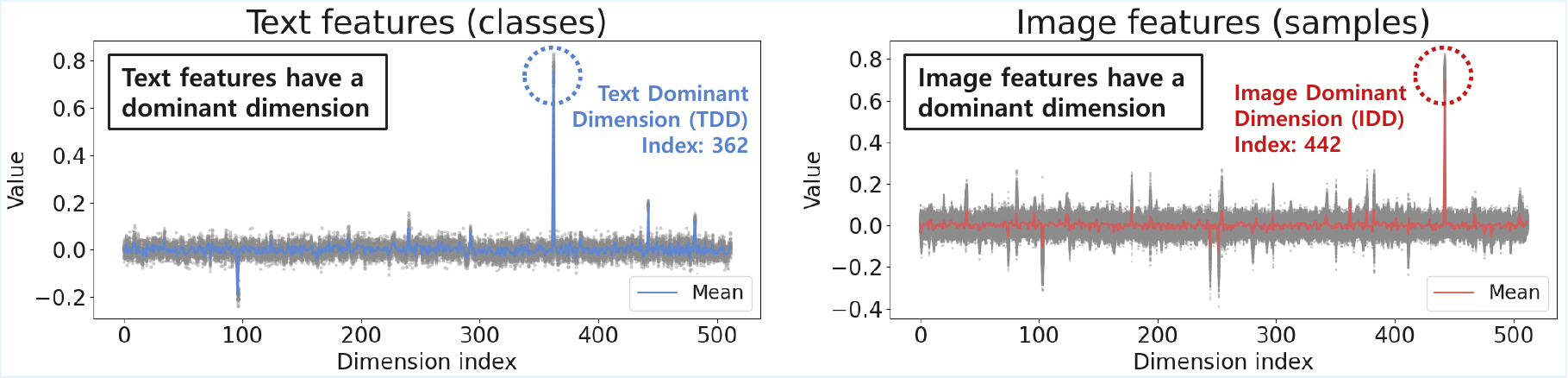}
    \caption{Comparison of feature values across dimensions for text and image features. In CLIP, the modality gap causes features from each modality to be positioned in different spaces, a phenomenon that manifests as the dominant influence of a few dimensions.}
    \label{fig:modalitygap}
\end{figure}

The correlation between average text feature dispersion (ATFD) and expected calibration error (ECE)~\citep{naeini2015obtainingece} has been demonstrated through various experiments~\citep{yoon2024ctpt}. Nevertheless, Figure~\ref{fig:ece_atfd} shows that calibration capability can be achieved despite low ATFD, suggesting that underlying causal factors remain to be uncovered. In particular, we conjecture that the modality gap~\citep{liang2022mind} between image and text features plays a crucial role in this phenomenon. A detailed discussion of our geometric intuition is provided in Section~\ref{geometry}.

Contrastive VLMs such as CLIP are trained by maximizing the cosine similarity between image and text embeddings. Nevertheless, a significant modality gap between image and text features is widely observed, and various approaches have been proposed to understand and exploit this phenomenon~\citep{ouali2023black,oh2023geodesic}. \textbf{The modality gap is not expressed uniformly across all feature dimensions; instead, it is concentrated in a single dominant dimension.} As depicted in Figure~\ref{fig:modalitygap}, modality gap arises from distinct bases in the image and text representations. These dominant dimensions are consistently observed across all text classes and all image samples. We denote the dimension prevailing in textual embeddings as the text-dominant dimension (TDD) and the corresponding one in visual embeddings as the image-dominant dimension (IDD).

A dominant dimension significantly influences the final logit computed via the inner product of image and text features and exhibits high sensitivity. To investigate the effects of TDD and IDD, we compared predictions in which the TDD and IDD dimensions were replaced with their respective mean values, thereby limiting the contribution of these dominant dimensions. Interestingly, despite restrictions on dominant dimensions, there are cases where both accuracy and ECE are improved. In particular, for TDD, the average ECE is improved for both zero-shot CLIP and TPT. Based on these observations, we propose \textbf{d}imensional entropy maximization for \textbf{t}est-time \textbf{p}rompt \textbf{t}uning (D-TPT), a calibration method that mitigates uncertainty by reducing dependence on dominant dimensions and maximizing the entropy of text features to better exploit the remaining dimensions.

We evaluate calibration performance across five calibration metrics on fine-grained classification datasets comprising eleven datasets and natural distribution shifts datasets, which are four ImageNet variants. Our method that regularizes the distribution in intra-text features provides competitive performance compared to existing methods that increase inter-text feature diversity.
We note that prior work has shown that the modality gap is primarily induced by a few dominant embedding dimensions~\citep{schrodi2024two}. Consistent with this, we independently observed the same phenomenon and confirm that such dominant dimensions are critical to prompt characteristics. Beyond this confirmation, our contribution lies in proposing methods that leverage this hidden property to improve calibration performance.

\section{Related Work}
\textbf{Prompt tuning.}
Pretrained models are utilized in various applications, and prompt tuning is a parameter-efficient fine-tuning approach that adapts large pretrained models to downstream tasks by training only a small number of input tokens~\citep{prompt_tuning}. Prompt tuning has become a standard solution alongside LoRA~\citep{hu2022lora} and adapters~\citep{rebuffi2017learning,zhang2021tip}, demonstrating its effectiveness across vision models~\citep{l2p,jia2022visual,bahng2022exploring} and language models~\citep{prefix_tuning,liu2021p}. In VLMs, CoOp~\citep{zhou2022coop} demonstrates that input contexts such as “a photo of a [class]” significantly influence downstream performance. To overcome the limitations of manual prompt engineering, CoOp introduced context optimization, where prompts are trained rather than hand-crafted. Building on this idea, prompt tuning for VLMs has since been widely explored. Notably, CoCoOp~\citep{zhou2022cocoop} utilizes image features to generate image-conditioned prompts that enhance generalization to unseen classes, while MaPLe~\citep{khattak2023maple} extends context tuning by applying prompts to both the image encoder and the text encoder.

\textbf{Test-time adaptation.}
Deep neural networks show strong performance in visual tasks but remain vulnerable under distribution shifts~\citep{domainshift}. Test-time adaptation has been proposed to address dynamically changing real-world environments by enabling pretrained models to adapt during inference in an unsupervised manner, thereby avoiding the need for additional pretraining. Research on test-time adaptation has primarily focused on entropy minimization~\citep{wang2021tent,niu2023sar,lee2024deyo,han2025ranked} and consistency regularization~\citep{Wangetal2022cotta,dobler2023rmt,liu2023vida}. Under this paradigm, TPT~\citep{shu2022tpt} adopts entropy minimization through augmented input views and prompt tuning, yielding effective adaptation of VLMs during inference and providing a foundation for subsequent research~\citep{feng2023diverse,zhang2024robust,xiao2025dynaprompt,sheng2025r}.

\textbf{Model calibration.}
Modern neural networks often suffer from overconfidence, which leads to miscalibration~\citep{guo2017calibration}. Calibration denotes the alignment between predicted probabilities and the true likelihood of correctness, and it is essential for reliable deployment in real-world settings. Existing methods include post-hoc calibration approaches~\citep{platt1999probabilistic,guo2017calibration,zhang2020mix} such as temperature scaling and regularization-based approaches~\citep{kumar2018trainable,mukhoti2020calibrating} that refine the objective function. Given the calibration problems of test-time adaptation in VLMs, C-TPT~\citep{yoon2024ctpt} improves diversity by encouraging text embeddings to disperse from the centroid. In contrast, SaLS~\citep{murugesan2024robust} argues that miscalibration originates from the logit range and proposes a method that adjusts logits within the zero-shot range. Furthermore, O-TPT~\citep{sharifdeen2025otpt} promotes orthogonality across text features, which enhances class separability and mitigates directional bias.

\section{Preliminary}
\textbf{Zero-shot classification.} CLIP~\citep{radford2021clip} is pretrained on massive image and text pairs to maximize the cosine similarity between the outputs of the image encoder $E_{visual}$ and the text encoder $E_{text}$. For zero-shot inference, the inputs to the image encoder and the text encoder are an image $x$ and a set of context prompts $\{[\mathrm{p}; y_c]\}_{c=1}^C$, where $\mathrm{p}$ denotes the prompt and $y_c$ denotes the class context, where $C$ corresponds to the number of classes. Following previous work~\citep{sharifdeen2025otpt}, we adopt the prompt ``a photo of a [class]'' as the base prompt. Accordingly, the predicted probability of class $c$ for an image $x_i$ is given by
$p_{i,c} = \frac{\exp (\tau \cdot \cos(t_c,v_i) )}{\sum_{k=1}^C \exp (\tau \cdot \cos(t_k,v_i) )},$
where $v_i = E_{visual}(x_i)$ is the image feature and $t_c = E_{text}([\mathrm{p}; y_c])$ is the text feature, and $\tau$ denotes the learnable logit scaling factor.

\begin{wrapfigure}{r}{0.5\textwidth}
\vspace{-4.5mm}
\setlength{\intextsep}{1pt}  
\setlength{\columnsep}{1pt}
\begin{algorithm}[H]
\caption{Pytorch-style pseudo-code of D-TPT}
\begin{algorithmic}
\State \textcolor{ForestGreen!70}{\# temperature: $\tau$, threshold: $\rho$}
\label{algorithm}
\State \textcolor{ForestGreen!70}{\# uniform distribution: U, hyperparameter: $\lambda$}
\State \textcolor{blue!50}{\# CLIP procedure}
\State $v = \text{image\_encoder}(\text{image})$
\State $t = \text{text\_encoder}(\text{text})$ 
\State $\text{logits} = \tau \ast \text{cosine\_similarity}(v,t)$ 

\State \textcolor{blue!50}{\# TPT procedure}
\State $p = \text{softmax}(\text{logits})$
\State idx = \text{select\_confident\_index}$(p,\rho)$ 
\State \text{tpt\_loss} = entropy(p[idx])

\State \textcolor{blue!50}{\# D-TPT procedure}
\State \text{dem\_loss} = kl\_divergence(softmax$(t)$,U) 
\State \text{total\_loss} = \text{tpt\_loss} + $\lambda \ \ast$ \text{dem\_loss}

\State total\_loss.backward()
\State optimizer.step()
\end{algorithmic}
\end{algorithm}
\vspace{-8mm}
\end{wrapfigure}

\textbf{Test-time prompt tuning protocol.} Following test-time prompt tuning approaches~\citep{shu2022tpt,yoon2024ctpt,sharifdeen2025otpt}, we aim to improve performance by updating the prompt $\mathrm{p}$ during inference on a single test sample $\{x_i^{test}, y_i^{test}\} \in \{\mathcal{X}_{test}, \mathcal{Y}_{test}\}$. Here, $\mathcal{Y}_{test}$ is only used for performance evaluation and is not accessed during test-time. During inference, the prompt is tuned in an unsupervised online manner. Specifically, TPT~\citep{shu2022tpt} generates additional $N-1$ augmented views for each test image, and optimizes the prompt by minimizing the entropy of high-confidence samples. The objective function is defined as
\begin{align}\label{eqn:tpt}
\mathcal{L}_{tpt}
= \frac{1}{\rho N} \sum_{i=1}^{N} \mathbb{I}[H(p_i) < \rho] \, H(p_i),
\end{align}
where $H$ denotes the entropy (i.e., $H(p_i) = - \sum_c p_{i,c} \log p_{i,c}$) is a confidence threshold, and $N$ is the number of inputs. 
Algorithm~\ref{algorithm} illustrates the overall procedure of test-time prompt tuning.
Based on the initial zero-shot predictions, TPT minimizes the entropy of high-confidence samples, while our proposed D-TPT further improves calibration through the regularization loss function described in Section~\ref{sec:method}. The prompt is updated via gradient descent on the objective function. After adaptation and prediction for a single test sample, the model is reset to its initial weights.

\textbf{Evaluation metrics.} 
We use five evaluation metrics in this study: four calibration metrics (ECE, AECE, MCE, and AURC) and classification accuracy. Expected calibration error (ECE)~\citep{naeini2015obtainingece} measures the average discrepancy between predicted confidence and accuracy, and the adaptive ECE (AECE)~\citep{mukhoti2020calibrating} addresses binning sensitivity by employing adaptive binning. Moreover, the maximum calibration error (MCE)~\citep{naeini2015obtainingece} captures the worst-case miscalibration across bins, and the area under the risk–coverage curve (AURC)~\citep{geifman2018biasaurc} quantifies how well a model’s confidence separates correct from incorrect predictions. For all calibration metrics, lower values indicate better calibration.

\section{Method}\label{sec:method}
\subsection{Dimensional sensitivity}
\begin{figure}[t]
  \centering
  \subcaptionbox{Sensitivity analysis.\label{fig:sensitivity}}{%
    \includegraphics[width=0.25\linewidth]{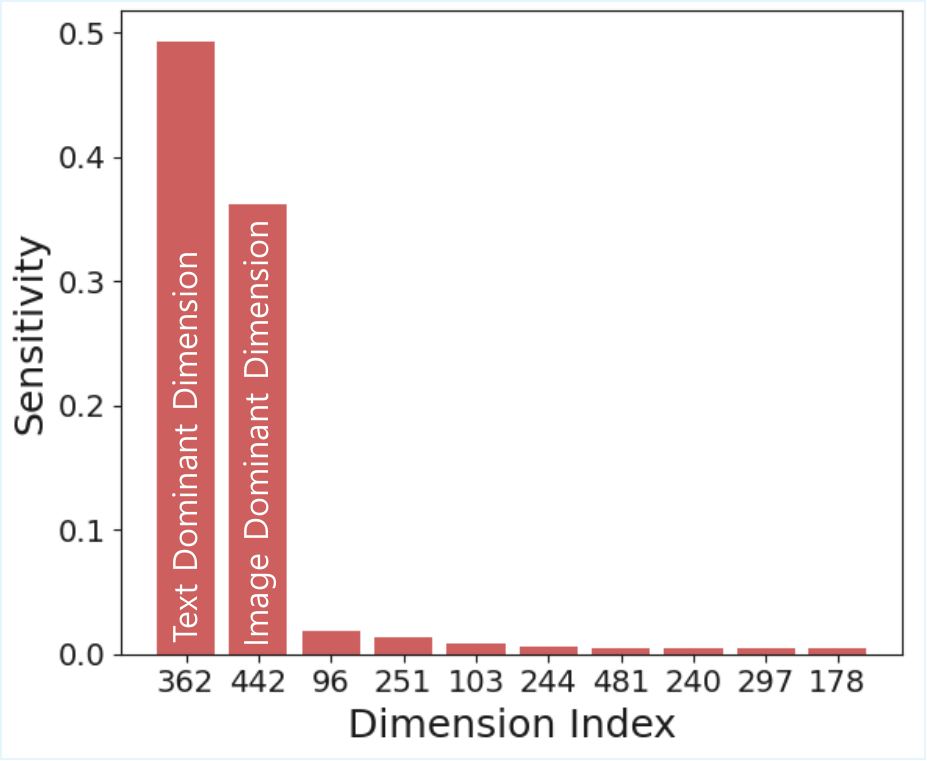}
  }
  \hfill
  \subcaptionbox{Effect of dominant dimensions.\label{fig:dominant}}{%
  \begin{minipage}{0.73\linewidth}
    \centering
    \vcenteredhbox{\includegraphics[width=0.34\linewidth]{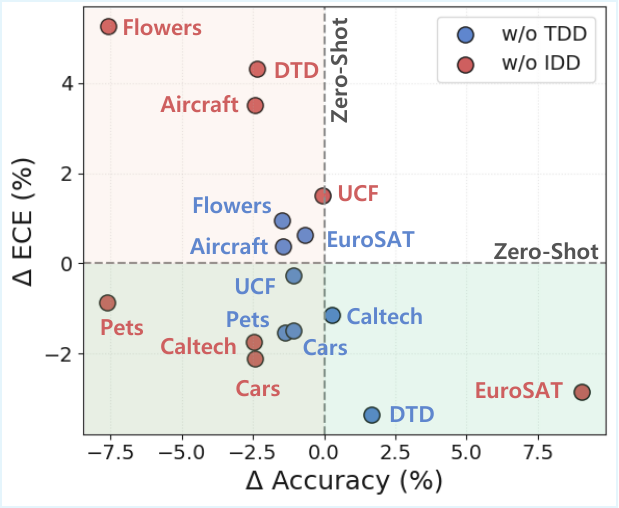}}
    \vcenteredhbox{%
      \setlength\tabcolsep{3pt}
      \resizebox{0.63\linewidth}{!}{%
        \begin{tabular}{l|cc|cc|cc}
          \toprule
          \multirow{2}{*}{Dataset} & \multicolumn{2}{c|}{TPT} &           
          \multicolumn{2}{c|}{\textcolor{brickblue}{\textbf{w/o TDD}}} & 
          \multicolumn{2}{c}{\textcolor{brickred}{\textbf{w/o IDD}}} \\          
           & Acc. $(\%)\uparrow$ & ECE $(\%)\downarrow$& Acc. $(\%)\uparrow$ &
           ECE $(\%)\downarrow$ & Acc. $(\%)\uparrow$ & ECE $(\%)\downarrow$ \\
          \midrule
          DTD      & 46.99 & 20.78 & 47.04 & 17.39 & 45.51 & 27.67 \\
          Flowers  & 69.06 & 13.29 & 68.86 & 8.91 & 64.11 & 13.56 \\
          Aircraft & 23.46 & 16.96 & 22.62 & 17.07 & 21.81 & 27.87 \\
          Pets     & 87.49 & 5.34 & 87.90 & 4.60 & 87.08 & 6.14 \\
          Caltech  & 94.32 & 4.70 & 94.12 & 3.57 & 93.96 & 3.27 \\
          UCF      & 67.94 & 11.72 & 68.20 & 10.22 & 66.93 & 16.88 \\
          EuroSAT  & 42.62 & 20.50 & 42.68 & 19.46 & 42.52 & 30.71 \\
          Cars     & 66.20 & 5.43 & 66.12 & 5.49 & 65.07 & 9.64 \\
          \midrule\midrule
          Mean     & 62.26 & 12.34 & 62.19 & 10.84 & 60.87 & 16.97 \\
          \bottomrule
        \end{tabular}
      }
    }
  \end{minipage}
}

  \caption{
    Analysis of the impact of dominant dimensions. \textbf{(a)} We present the top-10 values and their indices for sensitivity, which is the change in the prediction distribution when masking the values of each dimension. For both modalities, the dominant dimensions TDD and IDD significantly influence predictions. \textbf{(b)} Accuracy and ECE are reported with TDD and IDD replaced by their class-wise mean values. For zero-shot CLIP (left), replacing TDD leads to average improvements in ECE across datasets. For TPT (right), replacing TDD also yields a consistent reduction in mean ECE.
  }
  \label{fig:dim_masking}
\end{figure}

Given that CLIP obtains logits based on cosine similarity between image and text features, where the dominant dimension significantly contributes to prediction. In order to quantify the contribution of each feature dimension, we define the sensitivity of individual dimensions as
\begin{align}
    s_i = KLD(p_m||q),
\end{align}
where $q=\sigma(\tau \cdot \cos(t,v))$ denotes the original prediction probability, $\sigma(\cdot)$ denotes a softmax function, and $p_m$ denotes the prediction obtained by masking the $m\text{-}th$ dimension, $KLD(\cdot\mid\mid\cdot)$ denotes the Kullback–Leibler divergence. Figure~\ref{fig:sensitivity} presents the sensitivity analysis across feature dimensions, which shows that dominant dimensions exhibit substantially higher sensitivity. Moreover, Figure~\ref{fig:dominant} illustrates performance changes when dominant dimensions are replaced with their mean. For zero-shot CLIP, constraining these dimensions improves accuracy and calibration capability in some cases. For TDD, consistent reductions in average ECE were observed in both zero-shot CLIP and TPT. These results indicate that dominant dimensions have a pronounced impact on prediction, while their sensitivity can increase predictive uncertainty.

\subsection{D-TPT: Dimensional entropy maximization for test-time prompt tuning}

Building upon an analysis of dominant dimensionality and sensitivity, we introduce dimensional entropy maximization (DEM). We regularize the text feature distribution across all dimensions to be close to a uniform distribution, to reduce the effect of dominant dimensions and increase the contribution of hidden dimensions. Accordingly, our regularization loss term $\mathcal{L}_{DEM}$ is defined as

\begin{align}
\mathcal{L}_{DEM} = \frac{1}{C} \sum_{c=1}^C KLD(\sigma(\bar{t}_c)||U),
\end{align}
where $U$ denotes the uniform distribution and $\bar{t}_c$ denotes the normalized text feature corresponding to the $c$-th class. Using $\mathcal{L}_{DEM}$ as a regularization term for calibration, it can be easily applied to TPT via a plug-and-play approach. Hence, the total loss function of the proposed D-TPT is as follows:

\begin{align}
\mathcal{L}_{D-TPT} = \mathcal{L}_{TPT} + \lambda \cdot \mathcal{L}_{DEM},
\end{align}

where $\lambda$ is a hyperparameter. The prompts are updated by minimizing the overall loss function.

\subsection{Comparison with C-TPT and O-TPT}

\begin{figure}
    \centering
    \includegraphics[width=0.8\linewidth]{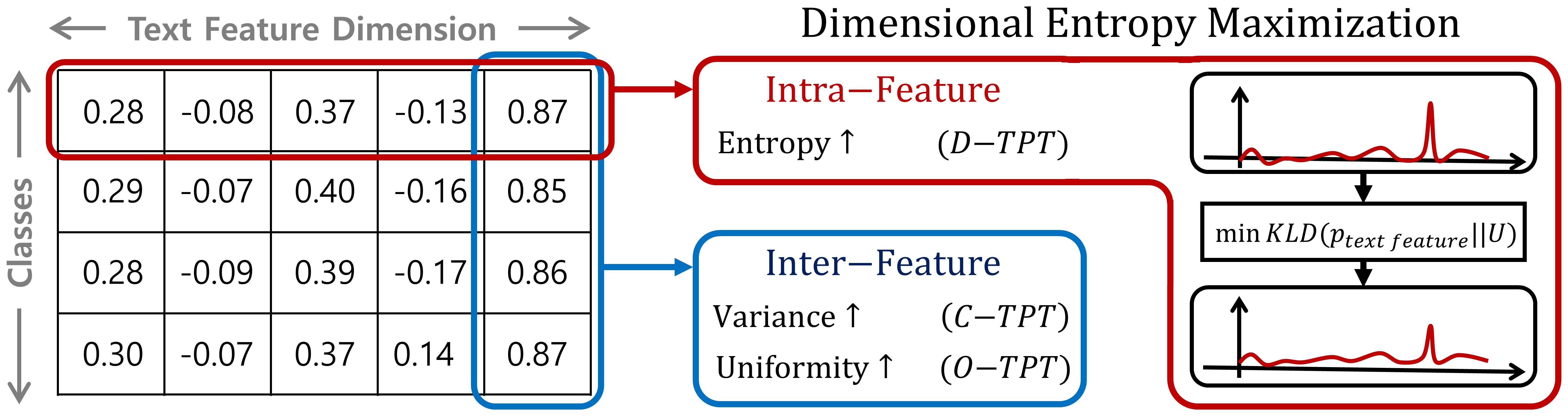}
    \caption{Conceptual illustration of D-TPT. Compared to inter-feature diversity based methods, we achieve improved calibration capability by regularization the distribution of intra-text features.}
    \label{fig:dtpt}
\end{figure}

C-TPT~\citep{yoon2024ctpt} and O-TPT~\citep{sharifdeen2025otpt} are calibration-focused variants of TPT~\citep{shu2022tpt} that aim to enhance the diversity of text features. Specifically, C-TPT induces text features to move away from the centroid, which increases the variance between text embedding,
\begin{align}
\mathcal{L}_{C-TPT} = \mathcal{L}_{TPT} + \lambda \cdot \frac{1}{C} \sum_{c=1}^{C} \biggl\lVert \bar{t}_c - \frac{1}{C}\sum_{j=1}^{C} \bar{t}_j \biggr\rVert_2.
\end{align}
Similarly, O-TPT maximizes the orthogonality between text features by minimizing the pairwise cosine similarity. Formally, the objective function of O-TPT is formulated as
\begin{align}
\mathcal{L}_{O-TPT} = \mathcal{L}_{TPT} + \lambda \cdot \lVert TT^{\top} - I_C \rVert^2_2,
\end{align}
where $T$ denotes the matrix of normalized text features, and $I_C$ is the $C$-dimensional identity matrix.
Figure~\ref{fig:dtpt} illustrates the comparison between our D-TPT and existing methods. The primary distinction is the direction of regularization. Previous methods focus on inter-feature structure across classes, while we focus on the intra-feature distribution. Despite this difference, both approaches improve calibration capability in practice. This difference naturally raises the question of why regularization strategies for text features contribute to improved calibration. While a complete theoretical framework remains an open challenge, we present a geometric analysis of the modality gap that provides a principled step toward addressing this question.

\subsection{Geometric support}\label{geometry}
\begin{figure}
    \centering
    \includegraphics[width=1.0\linewidth]{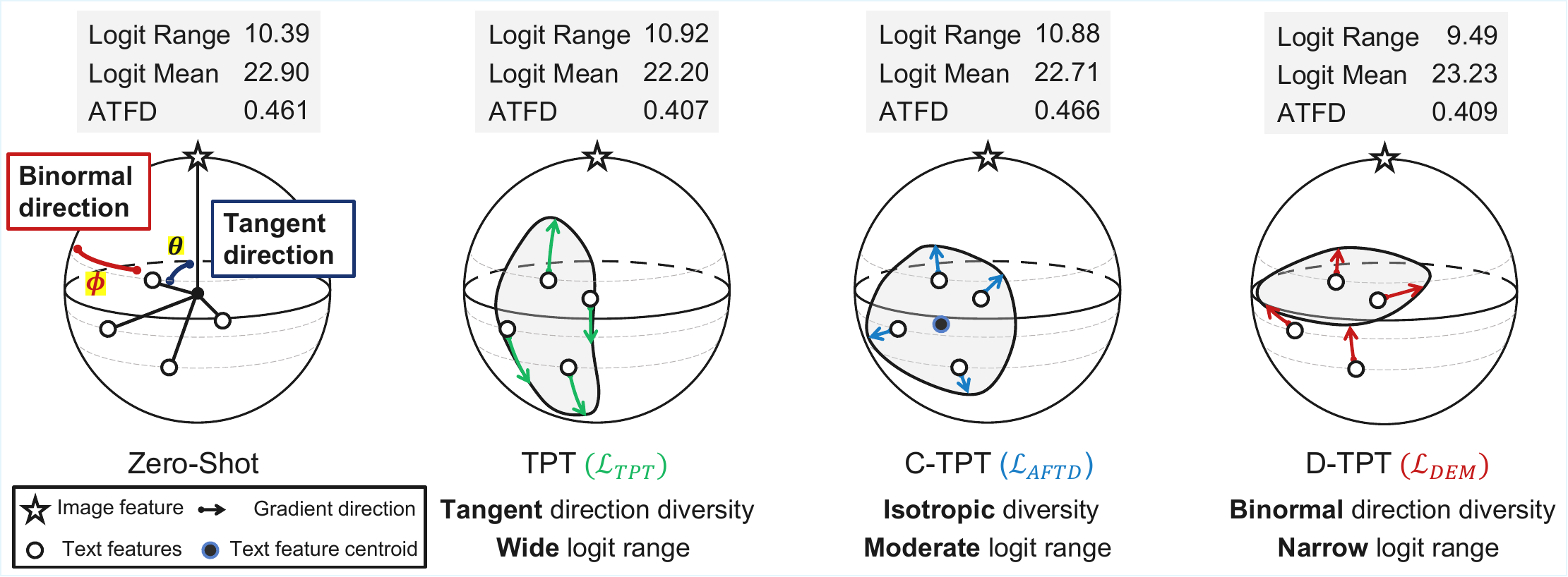}
    \caption{Geometric interpretation of prompt tuning methods on the hypersphere. On the hypersphere, the \textit{tangent direction} represents the geodesic from a text feature toward its corresponding image feature, the \textit{normal direction} corresponds to the radial axis from the center of the hypersphere to the feature, and the \textit{binormal direction} is defined as orthogonal to both tangent and normal directions. In this context, TPT aligns text features along the tangent direction of the hypersphere, which expands the logit range and leads to overconfidence. Meanwhile, C-TPT encourages outward displacement from the text centroid, thereby improving isotropic diversity. In contrast, proposed D-TPT maximizes entropy across feature dimensions, amplifying binormal components and reducing the modality gap.}
    \label{fig:logitrage}
\end{figure}

Figure~\ref{fig:logitrage} presents a geometric interpretation of the shift of text features under TPT, C-TPT, and proposed D-TPT.
In the single image case, the gradient of the TPT objective function is given by
\begin{align}
\frac{\partial \mathcal{L}_{tpt}}{\partial \theta} &= - \sum_{i=1}^{C} \left( \log p_{i,c} + 1 \right) p_{i,c} (1 - p_{i,c}) \frac{\partial z_{i,c}}{\partial \theta},
\end{align}
where $z_{i,c}$ and $p_{i,c}$ denote logit and probability corresponding to class $c$ for input $x_i$, respectively, and $\theta$ denotes the model parameters.

In TPT, the adoption of a confidence threshold ensures high-confidence prediction distributions. The prompt is optimized to increase the probability of the top-1 class while decreasing the probabilities assigned to the remaining classes. Since CLIP normalizes all feature vectors to unit length, the representations are constrained on a hypersphere. In this geometric structure, TPT encourages the text feature most similar to the image to move closer along the tangent direction of the hypersphere, while pushing other text features farther away. Consequently, this geometric effect enlarges the logit range and induces overconfidence.

In contrast, C-TPT optimizes text features to move away from the centroid. Let $k$ denote the number of text features that are closer to the image feature than the centroid, C-TPT increases the cosine similarity between the image feature and the top-$k$ text features. This mechanism mitigates overconfidence by compensating for the similarity reduction among non-top-1 classes. However, text features located farther from the centroid are explicitly assigned lower probabilities, which inherently carries the risk of increasing the logit range.

Our proposed D-TPT reduces the influence of dominant basis in the text features, thereby narrowing the modality gap. By maximizing entropy across all feature dimensions, D-TPT amplifies the binormal directions on the hypersphere. Empirically, our method exhibits similar ATFD to TPT but with a lower logit range, suggesting diversity in binormal directions. Since the logit mean is proportional to the average cosine similarity between image and text feature, we can confirm the reduction of the modality gap. Therefore, the reduced logit range of D-TPT leads to a decrease in softmax probabilities, which mitigates overconfidence~\citep{murugesan2024robust}.

\section{Experiments}
\subsection{Experimental setup}
\textbf{Datasets.}
Following the standard evaluation protocol for downstream tasks in VLMs~\citep{zhou2022coop,yoon2024ctpt}, we use datasets under natural domain shift, including ImageNet-A~\citep{hendrycks2021ina}, ImageNet-V2~\citep{recht2019inv2}, ImageNet-R~\citep{hendrycks2021inr}, and ImageNet-Sketch~\citep{wang2019ins}. For fine-grained classification, we use ImageNet~\citep{deng2009imagenet}, DTD~\citep{DTD}, Flowers102~\citep{Flower102}, Food101~\citep{Food101}, SUN397~\citep{SUN397}, FGVC-Aircraft~\citep{Aircraft}, Oxford Pets~\citep{Pets}, Caltech101~\citep{Caltech101}, UCF101~\citep{UCF101}, EuroSAT~\citep{eurosat}, and Stanford Cars~\citep{Cars}. Experimental results are reported as the average performance across all datasets for both the natural domain shift and fine-grained classification datasets. Moreover, we present the mean and standard deviation over three trials with random seeds 0, 1, and 2.

\textbf{Implementation details.}
We adopt TPT as the baseline and reproduce results under identical settings to enable further analysis of calibration metrics. Following TPT, each test sample is evaluated with a batch size of 64, comprising the original image and 63 augmented images. We use CLIP-ViT-B/16 and CLIP-RN50 as backbone architectures. Prompts are initialized with “a photo of a [class]” and optimized using AdamW~\citep{loshchilov2017decoupled} with a learning rate of 0.005. All experiments are conducted on NVIDIA RTX A6000 GPUs with 48 GB of memory. For practical applicability in test-time settings, we fix the hyperparameter $\lambda = 10^5$ across all datasets and architectures.

\begin{table}[t]
\centering
\small
\setlength{\tabcolsep}{5pt}
\scalebox{0.99}{
\begin{tabular}{l|ccccc}
\toprule
\multicolumn{6}{c}{\textbf{Fine-Grained Classification}}\\
\midrule
\multirow{2.5}{*}{Method}
& \multicolumn{5}{c}{CLIP-ViT-B/16}\\
\cmidrule(lr){2-6}
& Acc. $(\%)\uparrow$ & ECE $(\%)\downarrow$ & AECE $(\%)\downarrow$ & MCE $(\%)\downarrow$ & AURC $(\times10^3)\downarrow$ \\
\midrule
Zero Shot 
& 63.84 & 4.25 & 4.15 & 20.03 & 184.28 \\
TPT
& 65.09{\scriptsize $\pm$0.16} & 11.42{\scriptsize $\pm$0.19} & 11.37{\scriptsize $\pm$0.20} & 31.62{\scriptsize $\pm$5.02} & 185.64{\scriptsize $\pm$0.68}\\
C-TPT 
& 64.46{\scriptsize $\pm$0.15} & 4.97{\scriptsize $\pm$0.19} & 5.11{\scriptsize $\pm$0.16} & 21.54{\scriptsize $\pm$3.44} & 187.50{\scriptsize $\pm$0.46}\\
O-TPT 
& 63.98{\scriptsize $\pm$0.13} & 4.78{\scriptsize $\pm$0.13} & 4.88{\scriptsize $\pm$0.13} & 20.47{\scriptsize $\pm$5.55} & 187.25{\scriptsize $\pm$0.13} \\
\rowcolor{lightpink}{D-TPT (\textit{Ours})}
& 64.72{\scriptsize $\pm$0.07} & 4.18{\scriptsize $\pm$0.16} & 4.18{\scriptsize $\pm$0.19} & 21.18{\scriptsize $\pm$7.27} & 181.67{\scriptsize $\pm$0.30}\\
\midrule

& \multicolumn{5}{c}{CLIP-RN50}\\
\midrule
Zero Shot 
&56.05& 5.25 & 5.29 & 17.37 & 254.15  \\
TPT
& 58.07{\scriptsize $\pm$0.19} & 11.27{\scriptsize $\pm$0.18} & 11.26{\scriptsize $\pm$0.17} & 27.72{\scriptsize $\pm$3.87} & 251.15{\scriptsize $\pm$1.25}\\
C-TPT 
& 57.57{\scriptsize $\pm$0.12} & 6.20{\scriptsize $\pm$0.15} & 6.16{\scriptsize $\pm$0.18} & 23.96{\scriptsize $\pm$5.06} & 243.87{\scriptsize $\pm$0.50}\\
O-TPT 
& 57.36{\scriptsize $\pm$0.10} & 5.63{\scriptsize $\pm$0.16} & 5.68{\scriptsize $\pm$0.13} & 20.78{\scriptsize $\pm$3.78}& 244.96{\scriptsize $\pm$0.45} \\
\rowcolor{lightpink}{D-TPT (\textit{Ours})}
& 57.13{\scriptsize $\pm$0.12} & 5.91{\scriptsize $\pm$0.14} & 5.90{\scriptsize $\pm$0.13} & 22.44{\scriptsize $\pm$5.30} & 245.27{\scriptsize $\pm$1.00}\\
\bottomrule
\end{tabular}}
\caption{Calibration performance comparison between D-TPT and other baselines using CLIP-ViT-B/16 (above) and CLIP-RN50 (below) backbones on fine-grained classification datasets.} 
\label{tab:finegrained}
\end{table}

\begin{table}[t]
\centering
\small
\setlength{\tabcolsep}{5pt}
\scalebox{0.99}{
\begin{tabular}{l|ccccc}
\toprule
\multicolumn{6}{c}{\textbf{Natural Distribution Shifts}}\\
\midrule
\multirow{2.5}{*}{Method}
& \multicolumn{5}{c}{CLIP-ViT-B/16}\\
\cmidrule(lr){2-6}
& Acc. $(\%)\uparrow$ & ECE $(\%)\downarrow$ & AECE $(\%)\downarrow$ & MCE $(\%)\downarrow$ & AURC $(\times10^3)\downarrow$ \\
\midrule
Zero Shot 
& 57.19 & 4.93 & 4.90 & 10.54 & 216.23 \\
TPT
& 60.24{\scriptsize $\pm$0.10} & 11.77{\scriptsize $\pm$0.12} & 11.74{\scriptsize $\pm$0.10} & 21.72{\scriptsize $\pm$1.95} & 211.24{\scriptsize $\pm$0.62}\\
C-TPT 
& 58.33{\scriptsize $\pm$0.15} & 5.42{\scriptsize $\pm$0.07} & 5.45{\scriptsize $\pm$0.07} & 12.11{\scriptsize $\pm$1.07} & 210.52{\scriptsize $\pm$0.41}\\
O-TPT 
& 58.48{\scriptsize $\pm$0.07} & 5.14{\scriptsize $\pm$0.07} & 5.17{\scriptsize $\pm$0.14} & 11.33{\scriptsize $\pm$0.82} & 209.91{\scriptsize $\pm$0.15}\\
\rowcolor{lightpink}{D-TPT (\textit{Ours})}
& 57.87{\scriptsize $\pm$0.07} & 3.83{\scriptsize $\pm$0.04} & 3.85{\scriptsize $\pm$0.07} & 9.59{\scriptsize $\pm$1.26} & 211.31{\scriptsize $\pm$0.24}\\
\midrule

& \multicolumn{5}{c}{CLIP-RN50}\\
\midrule
Zero Shot 
& 40.66 & 7.19 & 7.13 & 18.66 & 395.77 \\
TPT
& 43.35{\scriptsize $\pm$0.09} & 17.06{\scriptsize $\pm$0.14} & 17.05{\scriptsize $\pm$0.15} & 32.21{\scriptsize $\pm$1.27} & 383.07{\scriptsize $\pm$0.84}\\
C-TPT 
& 41.76{\scriptsize $\pm$0.07} & 8.99{\scriptsize $\pm$0.12} & 8.97{\scriptsize $\pm$0.10} & 21.52{\scriptsize $\pm$0.53} & 390.19{\scriptsize $\pm$0.48}\\
O-TPT 
& 41.74{\scriptsize $\pm$0.09} & 8.97{\scriptsize $\pm$0.13} & 9.00{\scriptsize $\pm$0.14} & 21.36{\scriptsize $\pm$1.79} & 390.24{\scriptsize $\pm$0.80}\\
\rowcolor{lightpink}{D-TPT (\textit{Ours})}
& 42.92{\scriptsize $\pm$0.06} & 8.05{\scriptsize $\pm$0.09} & 8.07{\scriptsize $\pm$0.07} & 20.43{\scriptsize $\pm$1.01} & 379.73{\scriptsize $\pm$0.38}\\
\bottomrule
\end{tabular}}
\caption{Calibration performance comparison between D-TPT and other baselines using CLIP-ViT-B/16 (above) and CLIP-RN50 (below) backbones on natural distribution shifts datasets.} 
\label{tab:naturalshifts}
\end{table}

\subsection{Experimental results}
\textbf{Fine-grained classification.} Table~\ref{tab:finegrained} shows the average performance across eleven fine-grained datasets for evaluation on downstream tasks. Compared with zero-shot CLIP, TPT consistently shows improved accuracy but worse calibration performance.
Designed for improved calibration, C-TPT, O-TPT, and D-TPT apply additional regularization for calibration. These calibration-oriented methods achieve improved accuracy than zero-shot CLIP while reducing calibration error compared to TPT. Among them, D-TPT achieves 0.74\%, 0.60\%, and $5.58\times10^3$ improvements in accuracy, ECE, and AURC, respectively, compared to the previous state-of-the-art O-TPT on CLIP-ViT-B/16. On the other hand, O-TPT demonstrates the best performance in MCE, indicating the need for calibration evaluation across the various metrics. We do not achieve the best performance on CLIP-RN50, however, we achieve competitive performance between C-TPT and O-TPT. Detailed results are provided in~\ref{sec:detail_fg}.

\textbf{Natural distribution shifts.} Table~\ref{tab:naturalshifts} shows the average performance across four ImageNet variants, consisting of ImageNet-A/R/V2/Sketch for out-of-distribution evaluation. For CLIP-ViT-B/16, we achieve 0.61\% lower performance and 1.31\% improved calibration error compared to the previous state-of-the-art O-TPT. Compared to the zero-shot CLIP, we achieve improved accuracy and ECE by 0.68\% and 1.10\%, respectively, demonstrating the potential for improvement across all metrics through test-time prompt tuning. Furthermore, for CLIP-RN50, D-TPT demonstrates improved performance across all metrics compared to existing methods. Detailed results are provided in~\ref{sec:detail_ds}.

\subsection{Further analysis}
\begin{figure}
    \centering
    \includegraphics[width=1.0\linewidth]{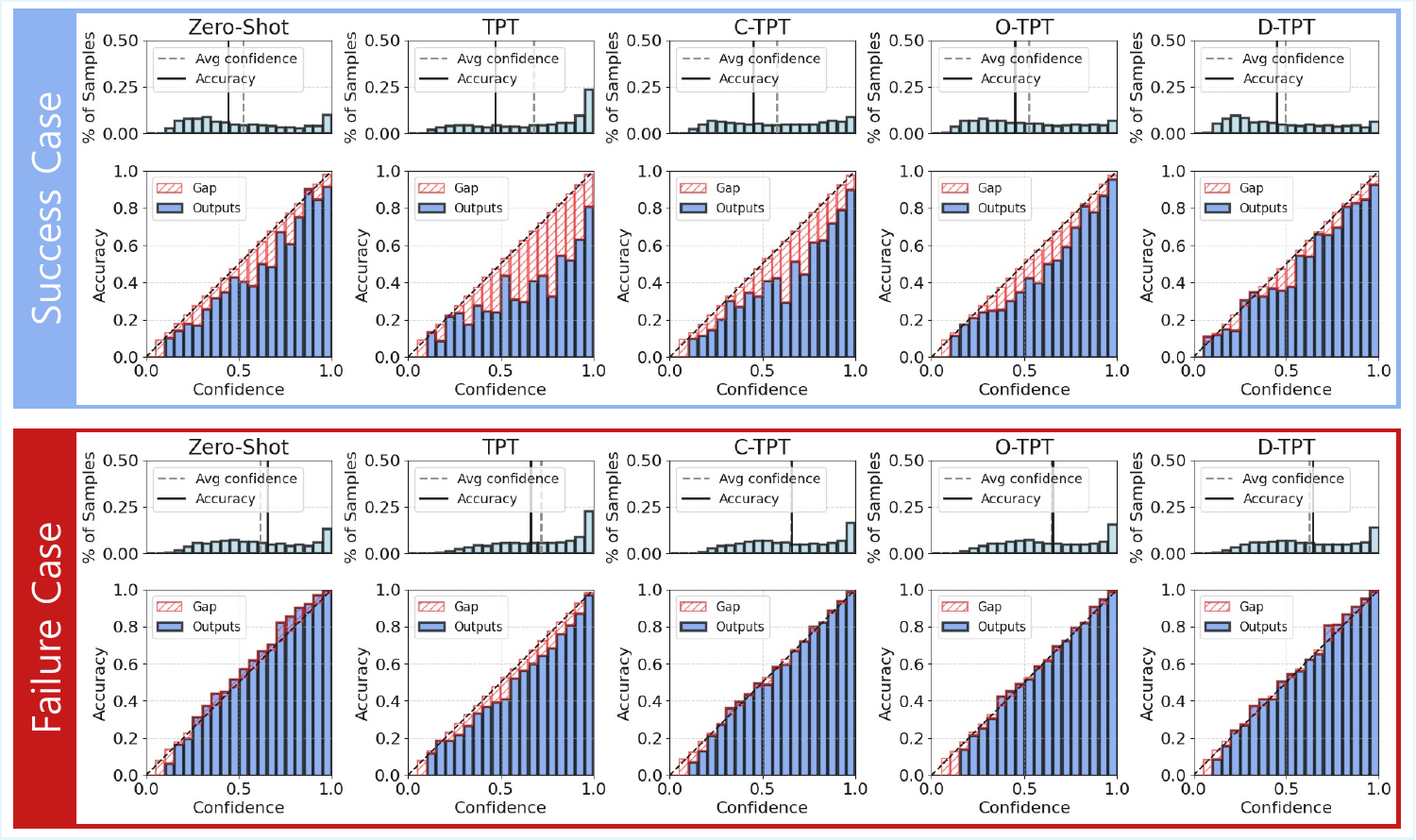}
    \caption{Confidence histograms and reliability diagrams. We present the sample distribution across confidence levels, as well as the gap between accuracy and confidence, for a successful case (DTD) and a failure case (Stanford Cars).}
    \label{fig:rd}
\end{figure}

\textbf{Failure case analysis.} Figure~\ref{fig:rd} presents confidence histograms (above) and reliability diagrams (below) to analyze the distribution of predictions and calibration trends with respect to confidence. Following C-TPT~\citep{yoon2024ctpt}, we report average accuracy and confidence over 20 bins. From both success and failure cases in comparison with existing methods, we observe that D-TPT more closely preserves the confidence distribution of the zero-shot CLIP. Therefore, the beneficial effects of D-TPT are particularly pronounced in cases where TPT significantly increases overconfidence. Furthermore, this observation is supported by the geometric interpretation discussed in Section~\ref{geometry}. Specifically, D-TPT aims to align all class text features with the image feature, hence mitigating changes in the features of non-top-1 classes induced by TPT. In contrast, C-TPT alleviates such changes primarily for classes that are closer to the centroid than the top-1 prediction. Consequently, our method can be understood as particularly effective in scenarios where TPT amplifies calibration error, since it counteracts the update directions for a broader set of class text features.

\begin{figure}
    \centering
    \includegraphics[width=0.8\linewidth]{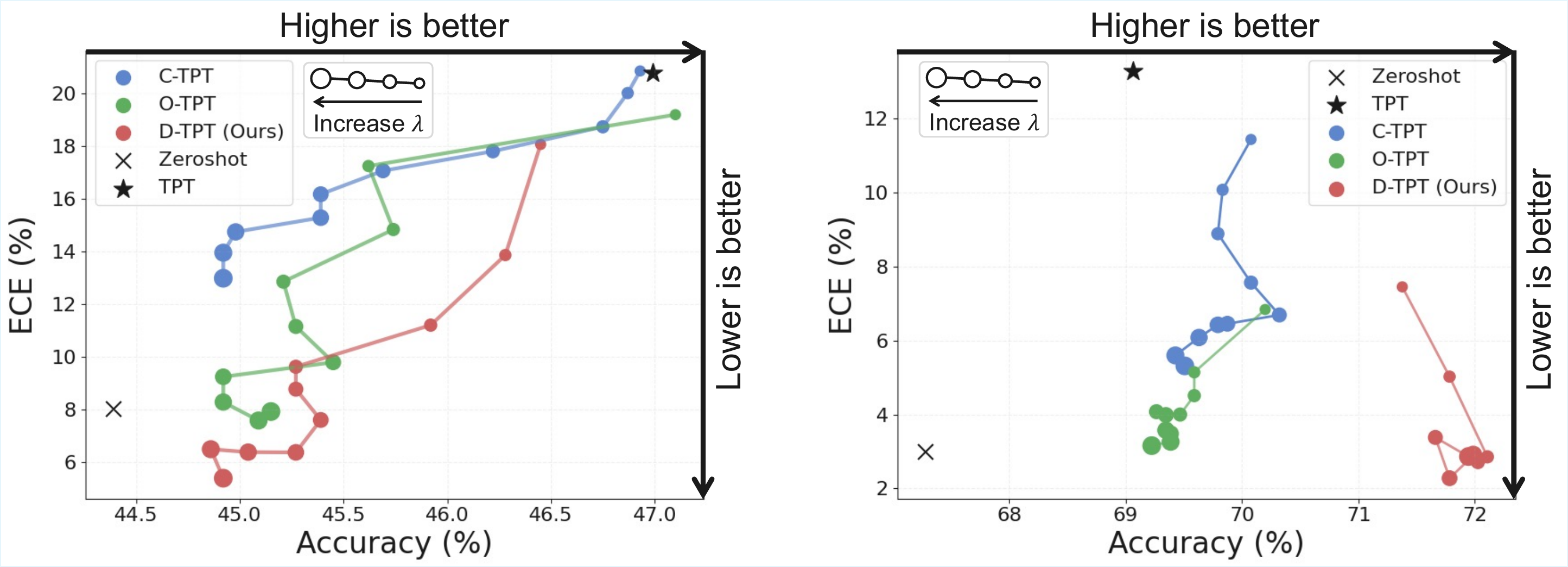}
    \caption{Pareto front analysis. We show the variation in ECE and Accuracy on DTD (left) and Flowers102 (right) as the hyperparameter $\lambda$ increases. Regarding the trade-off between Accuracy and ECE, a point closer to the right-under area indicates better performance.}
    \label{fig:pareto}
\end{figure}

\textbf{Pareto front analysis.}
Regularization methods for calibration improve calibration capability but often suffer from accuracy degradation~\citep{kumar2018trainable,karandikar2021soft,yoon2024ctpt}. Regarding the trade-off between accuracy and calibration performance, Figure~\ref{fig:pareto} presents the variations in performance of C-TPT, O-TPT, and D-TPT according to the coefficient of the regularization term. Increasing marker size indicates a larger influence of the regularization term with increasing hyperparameter $\lambda$. Experimental results demonstrate the superiority of the proposed method for a range of $\lambda$ values.

\textbf{Effect of prompt initialization.}
We present the performance of different prompt initialization schemes in Table~\ref{tab:resultcoop}. As described in the implementation details, we use the hand-crafted prompt “a photo of a [class]" as the initial prompt. Additionally, we conduct experiments using a prompt trained from CoOp~\citep{zhou2022coop} and another hand-crafted prompt, “a picture of a [class]”, to assess robustness to prompt initialization. In both cases, our method consistently improves ECE. Especially, under the initialization “a picture of a [class]”, our method achieves the best performance on most evaluation metrics.

\begin{table}[t]
\centering
\small
\setlength{\tabcolsep}{3pt}
\scalebox{0.95}{
\begin{tabular}{l|ccccc|ccccc}
\toprule
\multicolumn{11}{c}{\textbf{Fine-Grained Classification}}\\
\midrule
\multirow{2.5}{*}{Method}
& \multicolumn{5}{c|}{Trained prompt (CoOp~\citep{zhou2022coop})}& \multicolumn{5}{c}{Hand-crafted prompt (``a picture of a'')}\\
\cmidrule(lr){2-6}\cmidrule(lr){7-11}
& Acc. $\uparrow$ & ECE $\downarrow$ & AECE $\downarrow$ & MCE $\downarrow$ & AURC $\downarrow$ & Acc. $\uparrow$ & ECE $\downarrow$ & AECE $\downarrow$ & MCE $\downarrow$ & AURC $\downarrow$  \\
\midrule
CLIP-ViT-B/16 & 63.11 & 5.20 & 5.27 & 28.78 & 189.92 & 64.50 & 4.11& 4.16& 32.76& 178.13\\
TPT 
& 63.96 & 18.58 & 18.51 & 42.44 & 190.82 & 65.68& 10.86& 10.84& 31.89& 180.00\\
C-TPT 
& 64.04 & 9.95 & 9.90 & 29.51 & 188.13 & 65.23& 5.11& 5.19& 23.32& 177.90\\
O-TPT
& 63.60 & 7.80 & 7.73 & 27.97 & 187.97 & 64.59& 4.51& 4.50& 32.43& 179.07 \\
\rowcolor{lightpink}{D-TPT (\textit{Ours})}
& 63.20 & 6.96 & 6.96 & 24.43 & 195.41 & 65.22& 4.31& 4.37& 24.28& 171.27\\
\bottomrule
\end{tabular}}
\caption{Effect of initial prompt setting. We report accuracy (\%), ECE (\%), AECE (\%), MCE (\%) and AURC ($\times10^3$) for the trained prompt by CoOp and the hand-crafted prompt “a picture of a [class]”.}
\label{tab:resultcoop}
\end{table}

\section{Conclusion and Limitations}
In this paper, we introduce D-TPT, a simple yet effective solution to improve the calibration capability of test-time prompt tuning. Based on observations of dominant feature dimensions in contrastive VLMs, we demonstrate that constraining dominant features to specific dimensions can be beneficial in mitigating overconfidence. In contrast to diversity-based methods, our observations imply that useful hidden factors remain beyond diversity for improving the calibration of VLMs. Furthermore, we provide a geometric perspective on the modality gap to explain the success of regularization-based approaches for text features. Empirical results on standard VLM benchmarks confirm that D-TPT consistently improves calibration performance.

\textbf{Limitation and future work.} Nonetheless, our geometric interpretation is grounded in the dynamics of text features for a single image feature, leaving a practical gap in assumptions compared to approaches utilizing multiple image features via augmentation. Future work will aim to close the theoretical gap regarding the dynamics of text features and improvements in calibration performance, building a theoretical foundation based on the modality gap.

\bibliography{iclr2026_conference}
\bibliographystyle{iclr2026_conference}

\appendix
\section{Appendix}
\subsection{Detailed results on fine-grained classification}\label{sec:detail_fg}
Table~\ref{tab:full_vit} and~\ref{tab:full_rn} show the calibration performance for fine-grained classification across individual datasets. We report performance averaged over three different random seeds.

\subsection{Detailed results on natural distribution shifts datasets}\label{sec:detail_ds}
Table~\ref{tab:full_vit_ds} and~\ref{tab:full_rn_ds} present the dataset-wise average performance of CLIP-ViT-B/16 and CLIP-RN50 under natural distribution shifts, respectively.

\begin{table}[hbt!]
\centering
\small
\setlength{\tabcolsep}{4pt}
\scalebox{0.92}{
\begin{tabular}{l|ccccccccccc>{\columncolor{lightpurple}}c}
\toprule
Method & \rotatebox{90}{ImageNet} 
& \rotatebox{90}{DTD} 
& \rotatebox{90}{Flowers102} 
& \rotatebox{90}{Food101} 
& \rotatebox{90}{SUN397} 
& \rotatebox{90}{FGVC-Aircraft} 
& \rotatebox{90}{OxfordPets} 
& \rotatebox{90}{Caltech101} 
& \rotatebox{90}{UCF101} 
& \rotatebox{90}{EuroSAT}
& \rotatebox{90}{StanfordCars}
& \rotatebox{90}{Mean}\\
\midrule
\multicolumn{13}{c}{Accuracy $(\%)\uparrow$} \\
\midrule
Zero-Shot
&66.72& 44.33& 67.19& 83.66& 62.53& 23.73& 88.06& 93.23& 65.16& 42.06& 65.58& 63.84 \\
TPT &68.93&46.87&68.72&84.69&65.47&23.07&87.02&94.13&68.19&42.67&66.28&65.09 \\
C-TPT&68.10&45.31&69.24&83.09&64.50&23.93&88.28&93.41&65.03&42.41&65.70&64.46\\
O-TPT&67.30&44.86&69.08&82.76&63.26&23.42&88.19&93.25&63.83&42.60&65.21&63.98\\
\rowcolor{lightpink}{D-TPT}
&67.39& 44.96& 71.54& 83.11& 64.59& 23.94& 88.29& 93.09& 66.71& 44.02& 64.25& 64.72 \\
\midrule
\multicolumn{13}{c}{ECE $(\%)\downarrow$} \\
\midrule
Zero-Shot &1.91 &8.09 &3.19 &2.06 &2.10 &5.51 &4.34 &5.02 &2.93 &7.13 &4.50 &4.25 \\
TPT &10.53&21.25&13.55&4.33&11.25&17.16&5.81&4.51&11.49&20.37&5.33&11.42\\
C-TPT&3.10&12.75&5.48&3.34&5.02&4.39&1.81&3.90&2.25&11.22&1.46&4.97\\
O-TPT&1.94&8.14&3.56&4.15&8.45&3.85&2.41&3.95&2.90&11.43&1.80&4.78\\
\rowcolor{lightpink}{D-TPT} &2.73&5.20&2.82&4.07&4.50&3.74&2.38&5.43&2.65&9.78&2.62&4.18\\
\midrule
\multicolumn{13}{c}{AECE $(\%)\downarrow$} \\
\midrule
Zero-Shot
&1.90&7.82&2.55&2.01&2.02&5.74&4.38&5.01&2.79&7.02&4.44&4.15 \\
TPT&10.50&21.11&13.55&4.23&11.20&17.16&5.79&4.40&11.46&20.37&5.35&11.37 \\
C-TPT&3.04&12.89&5.74&3.46&5.07&4.90&1.86&3.96&2.45&11.29&1.53&5.11\\
O-TPT&1.91&8.12&3.88&4.35&8.45&4.50&2.36&4.04&2.67&11.44&2.00&4.88\\
\rowcolor{lightpink}{D-TPT}&2.73&5.21&2.98&4.28&4.50&3.96&2.25&5.45&2.56&9.63&2.42&4.18\\
\midrule
\multicolumn{13}{c}{MCE $(\%)\downarrow$} \\
\midrule
Zero-Shot &6.26&19.24&11.48&8.68&5.51&21.14&23.96&90.67&8.96&14.34&10.12&20.03 \\
TPT&19.05&40.79&29.93&8.71&19.88&45.69&45.62&74.13&23.88&29.71&10.47&31.62\\
C-TPT&7.98&26.61&16.75&8.88&12.56&14.61&27.69&72.82&18.21&22.39&8.44&21.54\\
O-TPT&11.10&22.79&15.27&9.77&29.33&13.35&18.55&58.70&10.37&23.33&12.63&20.47\\
\rowcolor{lightpink}{D-TPT}
&7.20&18.94&13.82&9.54&11.66&15.89&55.83&60.22&12.40&18.99&8.55&21.18 \\
\midrule
\multicolumn{13}{c}{AURC $(\times10^3)\downarrow$} \\
\midrule
Zero-Shot 
&138.50&313.46&112.32&54.83&182.54&577.84&21.07&12.37&122.47&355.43&136.23&184.28 \\
TPT&139.80&314.63&108.63&57.10&179.19&580.73&27.02&14.78&108.84&376.11&135.25&185.64\\
C-TPT&132.15&326.02&99.24&55.37&191.14&563.65&19.89&10.76&130.99&393.63&139.63&187.50\\
O-TPT&132.89&313.74&98.17&55.39&182.02&570.29&19.92&10.92&137.56&396.89&141.94&187.25\\
\rowcolor{lightpink}{D-TPT}
&135.37&317.81&93.06&51.91&164.52&563.30&19.82&11.75&114.45&378.85&147.57&181.67 \\
\bottomrule
\end{tabular}}
\caption{Full results on fine-grained classification using the CLIP-ViT-B/16 backbone.}
\label{tab:full_vit}
\end{table}

\begin{table}[t]
\centering
\small
\setlength{\tabcolsep}{4pt}
\scalebox{0.92}{
\begin{tabular}{l|ccccccccccc>{\columncolor{lightpurple}}c}
\toprule
Method & \rotatebox{90}{ImageNet} 
& \rotatebox{90}{DTD} 
& \rotatebox{90}{Flowers102} 
& \rotatebox{90}{Food101} 
& \rotatebox{90}{SUN397} 
& \rotatebox{90}{FGVC-Aircraft} 
& \rotatebox{90}{OxfordPets} 
& \rotatebox{90}{Caltech101} 
& \rotatebox{90}{UCF101} 
& \rotatebox{90}{EuroSAT}
& \rotatebox{90}{StanfordCars}
& \rotatebox{90}{Mean}\\
\midrule
\multicolumn{13}{c}{Accuracy $(\%)\uparrow$} \\
\midrule
Zero-Shot&58.13&40.37&61.59&73.93&58.73&15.81&83.59&86.21&58.84&23.70&55.60&56.05\\
TPT&60.74&41.86&62.47&74.98&61.43&17.57&84.58&87.41&60.79&28.51&58.42&58.07\\
C-TPT&60.05&41.78&64.70&74.81&60.95&16.60&83.43&87.17&60.26&27.24&56.32&57.57\\
O-TPT&58.99&41.84&65.57&74.62&59.62&17.06&83.17&86.77&59.93&27.66&55.70&57.36\\
\rowcolor{lightpink}{D-TPT}&60.39&42.40&60.29&74.24&59.80&16.93&83.13&86.99&58.72&30.30&55.27&57.13\\
\midrule
\multicolumn{13}{c}{ECE $(\%)\downarrow$} \\
\midrule
Zero-Shot&1.94&8.66&2.99&2.68&3.73&6.26&5.55&4.06&2.93&14.67&4.27&5.25\\
TPT
&11.38&25.40&13.82&5.17&9.03&15.64&3.80&4.20&10.87&20.83&3.84&11.27\\
C-TPT
&3.02&21.08&4.23&1.72&3.19&11.27&2.92&2.73&3.05&13.32&1.64&6.20\\
O-TPT&3.13&16.72&1.78&1.38&6.58&8.25&3.14&3.01&2.47&13.51&1.99&5.63\\
\rowcolor{lightpink}{D-TPT}
&2.68&13.53&4.72&5.10&4.02&7.79&5.29&5.25&2.44&6.14&8.04&5.91\\
\midrule
\multicolumn{13}{c}{AECE $(\%)\downarrow$} \\
\midrule
Zero-Shot &1.85&8.79&3.49&2.62&3.77&5.93&5.54&4.28&3.26&14.51&4.14&5.29\\
TPT &11.35&25.37&13.82&5.15&9.00&15.61&3.93&4.01&10.81&20.82&3.93&11.26\\
C-TPT&3.03&21.03&4.29&1.75&3.16&11.27&2.77&2.66&3.06&13.32&1.46&6.16\\
O-TPT&3.10&16.62&3.21&1.22&6.59&8.08&3.11&2.91&2.05&13.50&2.08&5.68\\
\rowcolor{lightpink}{D-TPT}&2.69&13.60&4.82&5.21&4.04&7.74&5.15&5.28&2.36&5.96&8.01&5.90\\
\midrule
\multicolumn{13}{c}{MCE $(\%)\downarrow$} \\
\midrule
Zero-Shot
&4.33&20.31&9.15&8.16&10.83&20.30&11.37&27.51&22.36&46.19&10.61&17.37\\
TPT&18.97&45.88&30.73&10.06&17.26&49.33&24.14&16.40&20.57&62.62&8.96&27.72\\
C-TPT&7.99&43.47&13.06&8.09&20.54&36.68&20.28&35.73&10.67&58.58&8.52&23.96\\
O-TPT&10.45&34.87&11.78&6.15&10.56&31.18&23.67&33.15&7.41&53.95&5.39&20.78\\
\rowcolor{lightpink}{D-TPT}
&6.21&31.62&13.07&9.15&16.17&35.65&20.54&31.93&8.24&59.77&14.49&22.44
\\
\midrule
\multicolumn{13}{c}{AURC $(\times10^3)\downarrow$} \\
\midrule
Zero-Shot&202.07&374.38&153.86&94.24&206.22&690.02&35.37&29.24&166.81&634.18&209.32&254.15
\\
TPT&195.57&371.36&160.31&94.53&202.81&667.79&34.31&36.17&170.28&631.39&198.15&251.15\\
C-TPT&192.87&363.66&135.67&90.08&198.30&677.65&36.89&27.69&159.00&591.52&209.21&243.87\\
O-TPT&196.42&362.02&130.74&90.84&199.86&674.67&38.73&28.42&163.93&594.92&213.97&244.96\\
\rowcolor{lightpink}{D-TPT}&194.21&372.90&161.13&93.94&210.55&675.49&38.76&30.45&176.89&528.28&215.39&245.27\\
\bottomrule
\end{tabular}}
\caption{Full results on fine-grained classification using the CLIP-RN50 backbone.}
\label{tab:full_rn}
\end{table}

\begin{table}[t]
\centering
\small
\setlength{\tabcolsep}{4pt}
\scalebox{0.92}{
\begin{tabular}{l|cccc>{\columncolor{lightpurple}}c}
\toprule
Method & \rotatebox{0}{ImageNet-A} 
& \rotatebox{0}{ImageNet-V2} 
& \rotatebox{0}{ImageNet-R} 
& \rotatebox{0}{ImageNet-Sketch} 
& \rotatebox{0}{Mean}\\
\midrule
\multicolumn{6}{c}{Accuracy $(\%)\uparrow$} \\
\midrule
Zero-Shot&47.80&60.85&73.98&46.12&57.19\\
TPT&53.15&62.90&76.95&47.97&60.24\\
C-TPT&49.41&61.75&74.85&47.31&58.33\\
O-TPT&49.91&61.67&75.22&47.11&58.48\\
\rowcolor{lightpink}{D-TPT}&49.12&61.35&74.05&46.97&57.87\\
\midrule
\multicolumn{6}{c}{ECE $(\%)\downarrow$} \\
\midrule
Zero-Shot&8.43&2.89&3.55&4.87&4.93\\
TPT&16.03&11.63&4.46&14.98&11.77\\
C-TPT&7.04&4.50&2.85&7.30&5.42\\
O-TPT&7.33&4.27&1.99&6.96&5.14\\
\rowcolor{lightpink}{D-TPT}&6.35&2.66&3.85&2.47&3.83\\
\midrule
\multicolumn{6}{c}{AECE $(\%)\downarrow$} \\
\midrule
Zero-Shot&8.40&2.74&3.59&4.87&4.90\\
TPT&15.94&11.63&4.43&14.97&11.74\\
C-TPT&7.15&4.49&2.87&7.30&5.45\\
O-TPT&7.43&4.22&2.05&6.96&5.17\\
\rowcolor{lightpink}{D-TPT}&6.38&2.70&3.88&2.45&3.85\\
\midrule
\multicolumn{6}{c}{MCE $(\%)\downarrow$} \\
\midrule
Zero-Shot&17.76&7.85&8.79&7.75&10.54\\
TPT&27.15&22.37&11.71&25.67&21.72\\
C-TPT&16.99&11.32&7.22&12.89&12.11\\
O-TPT&16.56&10.92&5.81&12.01&11.33\\
\rowcolor{lightpink}{D-TPT}&16.95&7.86&8.26&5.27&9.59\\
\midrule
\multicolumn{6}{c}{AURC $(\times10^3)\downarrow$} \\
\midrule
Zero-Shot&8.40&2.74&3.59&4.87&4.90\\
TPT&15.94&11.63&4.43&14.97&11.74\\
C-TPT&7.15&4.49&2.87&7.30&5.45\\
O-TPT&7.43&4.22&2.05&6.96&5.17\\
\rowcolor{lightpink}{D-TPT}&6.38&2.70&3.88&2.45&3.85\\
\bottomrule
\end{tabular}}
\caption{Full results on natural distribution shifts datasets using the CLIP-ViT-B/16 backbone.}
\label{tab:full_vit_ds}
\end{table}

\begin{table}[t]
\centering
\small
\setlength{\tabcolsep}{4pt}
\scalebox{0.92}{
\begin{tabular}{l|cccc>{\columncolor{lightpurple}}c}
\toprule
Method & \rotatebox{0}{ImageNet-A} 
& \rotatebox{0}{ImageNet-V2} 
& \rotatebox{0}{ImageNet-R} 
& \rotatebox{0}{ImageNet-Sketch} 
& \rotatebox{0}{Mean}\\
\midrule
\multicolumn{6}{c}{Accuracy $(\%)\uparrow$} \\
\midrule
Zero-Shot
&21.80&51.30&56.17&33.35&40.66\\
TPT&24.90&54.18&59.20&35.14&43.35\\
C-TPT&22.35&53.44&56.95&34.29&41.76\\
O-TPT&22.72&52.99&57.26&33.98&41.74\\
\rowcolor{lightpink}{D-TPT}&24.01&53.78&59.21&34.69&42.92\\
\midrule
\multicolumn{6}{c}{ECE $(\%)\downarrow$} \\
\midrule
Zero-Shot
&21.24&3.38&0.96&3.19&7.19\\
TPT&31.50&13.55&9.35&13.86&17.06\\
C-TPT&22.77&5.15&1.42&6.60&8.99\\
O-TPT&24.30&3.64&2.68&5.25&8.97\\
\rowcolor{lightpink}{D-TPT}&19.93&4.12&3.66&4.50&8.05\\
\midrule
\multicolumn{6}{c}{AECE $(\%)\downarrow$} \\
\midrule
Zero-Shot
&21.24&3.07&1.03&3.19&7.13\\
TPT&31.50&13.53&9.32&13.86&17.05\\
C-TPT&22.77&5.13&1.39&6.60&8.97\\
O-TPT&24.30&3.81&2.64&5.25&9.00\\
\rowcolor{lightpink}{D-TPT}&19.91&4.22&3.66&4.50&8.07\\
\midrule
\multicolumn{6}{c}{MCE $(\%)\downarrow$} \\
\midrule
Zero-Shot
&53.79&9.22&4.61&7.03&18.66
\\
TPT&59.11&25.02&18.65&26.06&32.21\\
C-TPT&54.57&12.96&4.25&14.31&21.52\\
O-TPT&54.80&10.16&8.20&12.26&21.36\\
\rowcolor{lightpink}{D-TPT}&50.48&11.47&7.37&12.41&20.43\\
\midrule
\multicolumn{6}{c}{AURC $(\times10^3)\downarrow$} \\
\midrule
Zero-Shot &695.94&259.96&188.28&438.88&395.77\\
TPT&672.47&251.35&179.22&429.27&383.07\\
C-TPT&690.08&248.86&188.44&433.38&390.19\\
O-TPT&687.95&248.85&188.70&435.47&390.24\\
\rowcolor{lightpink}{D-TPT}&667.63&248.11&173.44&429.73&379.73\\
\bottomrule
\end{tabular}}
\caption{Full results on natural distribution shifts datasets using the CLIP-RN50 backbone.}
\label{tab:full_rn_ds}
\end{table}

\end{document}